# Advancing rail safety: An onboard measurement system of rolling stock wheel flange wear based on dynamic machine learning algorithms

Celestin Nkundineza[1,4], James Ndodana Njaji[2], Samrawit Abubeker[2,3], Omar Gatera[4], Damien Hanyurwimfura[4]

[1]Department of Mechanical and Energy Engineering,
College of Science and Technology,
University of Rwanda,
Kigali, Rwanda
Email: c.nkundineza@ur.ac.rw

[2]African Railway Center of Excellence
Addis Ababa University,
Addis Ababa, Ethiopia, 1000

[3]Safety and Security Department,
Addis Ababa Light Rail Transit Service,
Addis Ababa, Ethiopia

[4]Africa Center of Excellence in Internet of Things
College of Science and Technology,
University of Rwanda,
Kigali, Rwanda

# ABSTRACT

Rail and wheel interaction functionality is pivotal to the railway system safety, requiring accurate measurement systems for optimal safety monitoring operation. This paper introduces an innovative onboard measurement system for monitoring wheel flange wear depth, utilizing displacement and temperature sensors. Laboratory experiments are conducted to emulate wheel flange wear depth and surrounding temperature fluctuations in different periods of time. Employing collected data, the training of machine learning algorithms that are based on regression models, is dynamically automated. Further experimentation results, using standards procedures, validate the system's efficacy. To enhance accuracy, an infinite impulse response (IIR) filter that mitigates vehicle dynamics and sensor noise is designed. Filter parameters were computed based on specifications derived from a Fast Fourier Transform analysis of locomotive simulations and emulation experiments data. The results show that the dynamic machine learning algorithm effectively counter sensor nonlinear response to temperature effects, achieving an accuracy of 96.5%, with a minimal runtime. The real-time noise reduction via IIR filter enhances the accuracy up to 98.2%. Integrated with railway communication embedded systems such as Internet of Things devices, this advanced monitoring system offers unparalleled real-time insights into wheel flange wear and track irregular conditions that cause it, ensuring heightened safety and efficiency in railway systems operations.

# 1. INTRODUCTION

Over years, railway industries have continuously sought ways to improve safety operation cost-effectiveness by finding better, efficient, effective technologies as solutions. Rail and wheel interactions are among the most components that demand safety measures in railroad transportation systems (1). This interaction results in wear of flange and wheel tread. Wear of the wheel flanges occurs when the wheel flange of the moving train wheel comes in contact with the rail. It happens when a railway vehicle moves on curved tracks or, irregular tracks, or when it undergoes hunting motion, realizes tractive and braking efforts, or when a vehicle is moving in high-speed mode. A very worn-out wheel flange means increased probability of train derailment, loss of life and economic damages. For reliability of the trains, it is important to monitor the wheel flange before it is reduced to the allowable safe minimum value. This monitoring allows train operator to schedule the maintenance efficiently and economically. In literature, there are various established methods to monitor the factors of wheel flange wear. These methods are based on monitoring wheel rail contact forces, derailment coefficient, and wheel profile and dimensions measurement. The main measurement techniques used in these methods include optical sensors, laser cameras, ultrasonic techniques, and strain gauges.

The movement of the rail vehicle is dictated by tractive effort and resistive forces. The result is complex dynamics of the rail vehicle- track system, including frictional forces between the wheel and rail, brake shoes and the wheel, and between other contacting parts. These frictional forces result in the release of heat energy within the contact zone, consequently leading to an increase in temperature(2). The temperature ranges are contingent upon various operational parameters of the rail vehicle, such as speed, duration of operation, and braking systems. In the case of light rail transit electric multiple units (EMU), temperatures typically remain within a lower range, typically not exceeding 60°C (as per data collected from the Addis Ababa Light Rail). This is attributed to the slow speeds at which these vehicles operate, frequent stops, and the utilization of disk-shoe brakes mounted on the axle. Consequently, the braking method exerts minimal influence on temperature elevation in the wheel.

Conversely, mainline trains and commuter trains and other types of trains that use wheel tread braking, experience significantly higher temperature ranges in the wheel contact zone. This is due to factors related to their operation over long distances without frequent stops, as well as the utilization of wheel-mounted shoe brakes. Commuter trains and metro trains that use wheel tread braking can experience temperatures ranging up to 120°C(3). Main-line heavy-haul train wheel can experience temperatures ranging up to 200°C(4). Consequently, sensors employed to measure wheel flange wear on locomotives operating on mainline railways may require distinct calibration and data acquisition methodologies compared to those used for light rail transit vehicles. This differentiation is particularly crucial if the sensors are sensitive to temperature fluctuations.

There are existing flange wear depth measurement methods, classified into contact and non-contact methods. The contact methods involve manual measurement techniques using handheld tool devices (5) and using electronic wheel profile meters when the rail vehicle is at the depot (6). Non-contact measurement methods are into two categories trackside and onboard measurement devices. The trackside electronic devices, ultrasonic sensors, lasers, and image processing devices process real-time measurements only when the train passes by them (7–10). The limitation of the trackside devices is that at different locations, there are no measurements recorded. It is disadvantaged and expensive to install a continuous trackside measurement device to monitor the wheel flange wear in terms of efficiency, and reliability. The onboard measurement devices include lasers, ultrasonic, and inductive devices (11–13). When contrasting the computer vision-based systems, which rely on cameras, laser sensors, and intricate image processing, with the alternative onboard method utilizing inductive sensing technology, a deeper analysis reveals more than just cost differentials. While the former may incur higher expenses due to equipment on track installation, they are less subjected to stresses of train operation and therefore have higher durability than the later. On the other hand, the onboard approach stands out for its simplicity, affordability, real time capability and impressive accuracy. For the computer vision-based systems with cameras, laser sensors, and image processing technologies taking flange measurements while the train is in operation, there is a cost-effect in equipment and installation along the track. The developed onboard flange measurements method based on inductive sensing technology in the other case is simple, affordable, with high accuracy and precision (13).

A notable limitation lies in the oversight of temperature effects resulting from heat transfer through radiation around the contact surface surroundings, an aspect not addressed in prior studies. Consequently, the efficacy of such a device would be optimized for rail vehicles that operate within stable temperature conditions, such as trams and metro systems. These modes of transportation typically maintain lower speeds and employ a braking system consisting of a disk and brake pads on the axle.

Conversely, mainline trains operate at higher speeds, run long distances before stops, and predominantly rely on brake shoes attached to the wheels, resulting in elevated wheel temperatures. This poses a considerable obstacle to optimizing the performance of the aforementioned device. Furthermore, it's worth noting that in the case of inductive sensor-based wheel flange wear measurement, noise removal was conducted offline, which could potentially limit the efficiency of real-time monitoring.

As a branch of artificial intelligence, machine learning uses various algorithms that are trained to predict responses or outputs of the system fed by certain inputs, which are similar to same population that the learning algorithm was trained with (14). The level of intelligence used currently in railway condition monitoring include integration of sensors, wireless sensor networks(15), internet of things(16,17), integration of simulation techniques to generate training data in some cases, and application of polynomial and logistic regression algorithms, Artificial Neural Networks, Support Vector Machines (18), and Deep Learning algorithms for predictive analytics (19,20), among others. However, for time varying sensor systems, in which parameters change over time, the inputs to the learning algorithm will differ, time by time, from population data that the system was trained with. To overcome this challenge, existing machine learning techniques in railway system condition monitoring uses deep learning algorithms such as Long-Short Term Memory (LSTM) (21). However, this technique is computationally intensive during its training because on one hand there require implicit method for model training, and on the other hand they require a large dataset feature (for example, the number of pixels in one picture) for model training resulting in large model parameters to be trained.

In this study, we collect training data from both inductive and temperature sensors (type K thermocouples) in relation to disk displacement (flange wear). We develop a decision key algorithm based on a regression model, which dynamically updates as new training data are incorporated into the structured database, accommodating variations in sensor parameters over time. Our initial experiments have revealed that the system's electrical parameters evolve with time, resulting in varying responses to identical temperature and displacement inputs. From collected data at Addis Ababa Light Rail Transit Service and Ethio-Djibouti Railway, the temperature at the wheel of EMU at a light rail does not exceed 60°C while the temperature of the locomotive wheel at the main line can rises up to 90°C. Traditional methods typically entail gathering new data and retraining the system, which can be cumbersome and disrupt the operation of rail vehicles due to the need for extensive data collection. Moreover, the conventional approach of retraining machine learning models involves numerical optimization algorithms, such as the widely-used gradient descent method. However, this process results in elevated computational runtime and necessitates rigorous validation for accuracy before deployment in the measurement system.

Our proposed self-updating algorithm aggregates data collected over time from manual measurements, enabling automatic updates to regression model coefficients through an explicit computation and iterations to determine orders of the polynomial features. These updates ensure correlation coefficients remain within a confidence interval of at least 95%. Notably, our method benefits from closed-form optimization matrix equations, significantly reducing computational runtime. Performance evaluation relies on achieving a minimum 95% confidence level in prediction accuracy, effectively mitigating bias issues. To address overfitting concerns, we employ a substantial training dataset at initial stage. This approach guarantees accurate prediction of flange wear based on sensor voltage and temperature inputs. In addition to the self-updating learning algorithm, real-time filtering effectively mitigates the impact of vibrations and signal noise, providing operators with timely insights into the status of the vehicle's wheel flange. We opt for an IIR filter due to its advantages of minimal delay, low memory requirements, and high resolution at low frequencies.

The proposed self-updating algorithms takes aggregated data overtime, which were taken by performing manual measurements. At any time, it is able to automatically update the regression model coefficients that achieve the correlation coefficient which is within the confidence interval (above 95%). Thus, it predicts the clearance between the sensor and the wheel flange accurately when it is fed by the sensor voltage and temperature. However, the self-updating algorithm does not take into account effects of vibrations and signal noises, and these effects must be filtered in real-time if the operators must be informed on the status of the vehicle wheel flange at any time. Therefore, an IIR filter is chosen for this task because of its advantage of less delay, less memory requirement and high resolution at low frequencies.

Over the past years, there are numerous researches on real-time software and hardware implementation for IIR digital filters. Woods et al. (22) designed a high-performance IIR digital filter chip utilizing circuit architecture based on the most significant bit-first arithmetic, demonstrating significant potential for full-scale development. Delta operators realizing digital filters gained interest due to their excellent finite word length performance under sampling rate. Kauraniemi et al. (23) focused on round off noise analysis, where the direct form II transposed delta structure showed the lowest quantization noise levels at its output. There was excellent noise filtering, but it came with additional implementation complexity compared to delay realization. Dehner (24) gave design procedures for cascades with second-order sections (SOS) indirect form and state-space form. Pairing and ordering of poles and zeros within the cascades were determined using dynamic programming.

Rabiul Islam et al. (25) proposed an architecture of a programmable digital IIR filter based on the XILINX FPGA board. Gate-level design analyzes the impulse response of the filter. It is disadvantaged when realizing higher-order filters as the speed, cost, and flexibility are affected by complex computations. Chen and Wang (26) presented a simple and more efficient method for IIR digital filter design based on Butterworth analog filter using the impulse invariant method. The design process involved these steps: designing a non-quantized IIR filter using MATLAB; converting the designed IIR filter rom M language to hardware description language (HDL) using AcceIDSP; finally verifying the design through timing simulation and synthesis using Xilinx ISE 10.1. This design method by far overcomes the ambiguity of IIR filter design with hardware description language directly. Hence, a cheaper, higher sampling rate, and real-time IIR design with AcceIDSP. Hourani et al. (27) utilized concepts inherent to both second-order sections (SOS) and state-space models to implement IIR filters in hardware. They compared the performance of the state space hardware model to the conventional IIR filter implementation using Spartan 3 XC3S400-4FG320 FPGA and a two-fold increase in hardware throughput using the state space SOS models was observed. Toledo-Pérez et al. (28) filtered noises using an IIR digital filter FPGA for Myoelectric signals.

Previous research on using inductive and temperature sensor for real-time flange measurements, studied effects of temperature on the inductive sensor, from 20°C − 73°C, and data fusion techniques were used to eliminate measurement drift (29). Noise in the measurements was filtered in offline data, measurements were taken to under temperatures up to 73°C, and noises from vibrations of the train during operations were not considered. This paper presents measurement data taken at temperatures up to 98°C. A self-updating agent is implemented to improve the accuracy of the system through optimization, reducing human interaction. Multi-body simulations of the locomotive are carried out. Frequency analysis of the results from prototype measurements and locomotive dynamics simulation is done to extract filter specifications. Finally, an online digital infinite impulse response filter is simulated in MATLAB/ Simulink to remove all noises from the system and train vibrations.

2. **METHODS OF DATA COLLECTION AND MACHINE LEARNING ALGORITHM DEVELOPMENT**

The work presented in this article consists on measuring the wheel flange wear depth in terms of the clearance between the sensor tip and wheel flange surface (see Figure 1). This work has followed and anticipated a methodology that is presented in Figure 2. On one hand, an experimental setup that emulates actual wheel flange wear depth measurement using inductive and temperature sensors is developed using a laterally translating disk, heat source, inductive displacement sensor that senses the lateral displacement of the disk ('Distance" or "Clearance"), data acquisition equipment, a computer and data acquisition software, LABVIEW (see Figure 3).

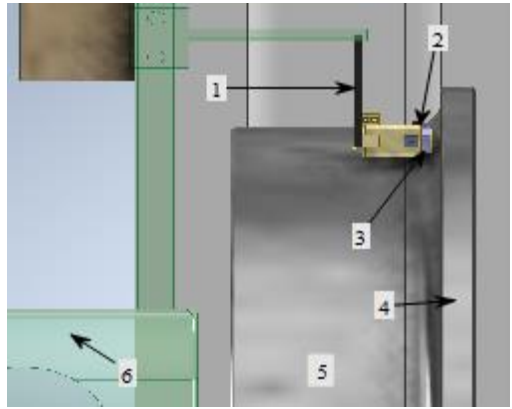

Figure 1. The sensor support structure and the rail wheel. (1) Sensor holding fixture, (2) thermocouple sensor, (3) inductive sensor, (4) wheel flange, (5) wheel tread, (6) bogie frame

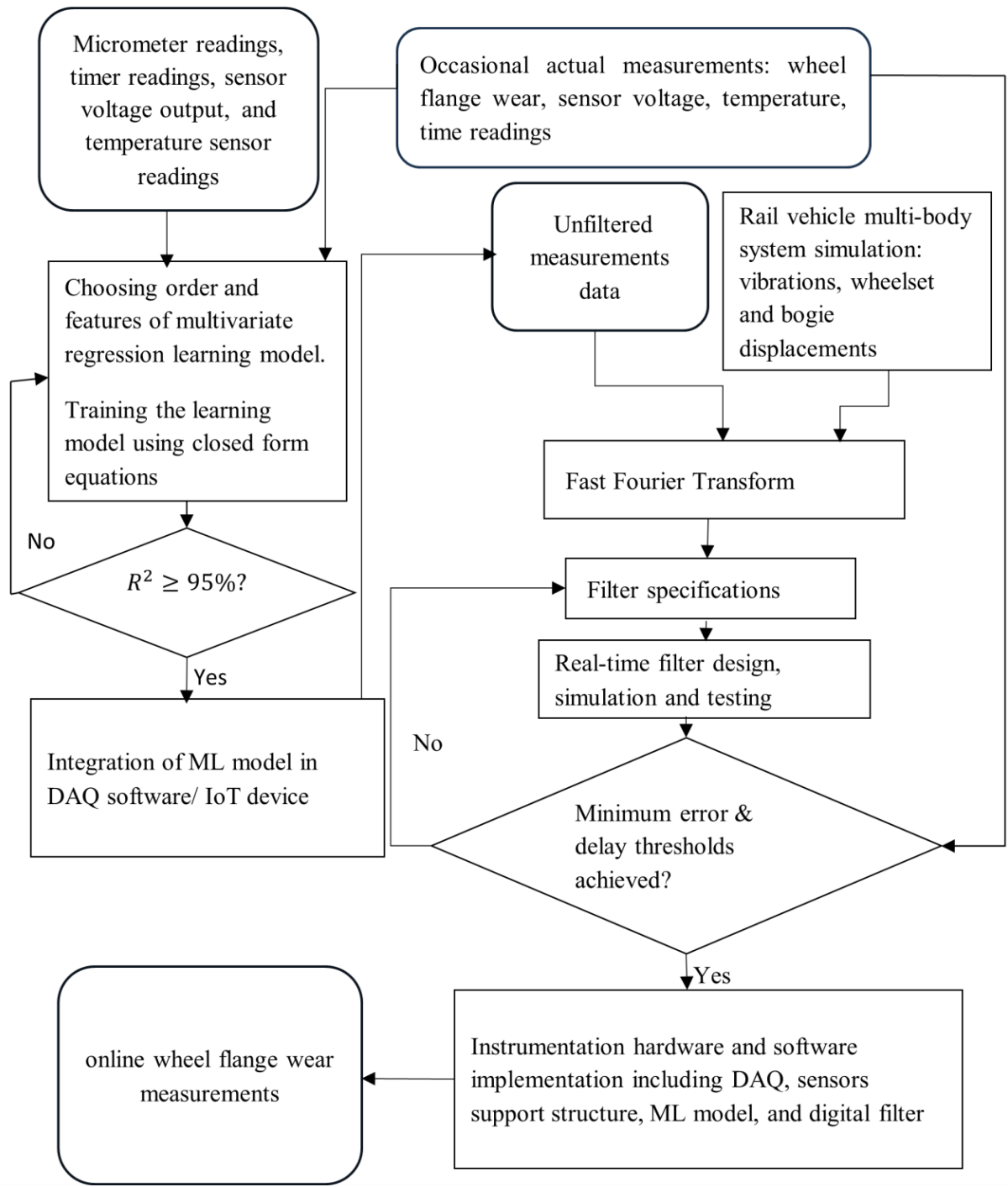

Figure 2. Information flow of research methods

### 2.1. Experiment setup and data collection

For data collection and testing of the system, a prototype for emulating the measurement system is set up (see Figure 3). The setup involves two-stage processes. Stage 1 involves taking circular disk input displacements along with inductive sensor voltage outputs, and surrounding temperature measurements, respectively using a micrometer, voltmeter, and thermocouple. These readings are considered as reference data with fewer systematic errors. Repeated similar multiple measurements are taken daily at different temperatures 19.8°C to 98°C. Stage 2 involves data acquisition to automatically measure data from the inductive sensor and thermocouple, writing them to a computer file. The LabVIEW software, is used with the data acquisition interface, and a NI USB 6221 board are used to record the sensor readings (29) .

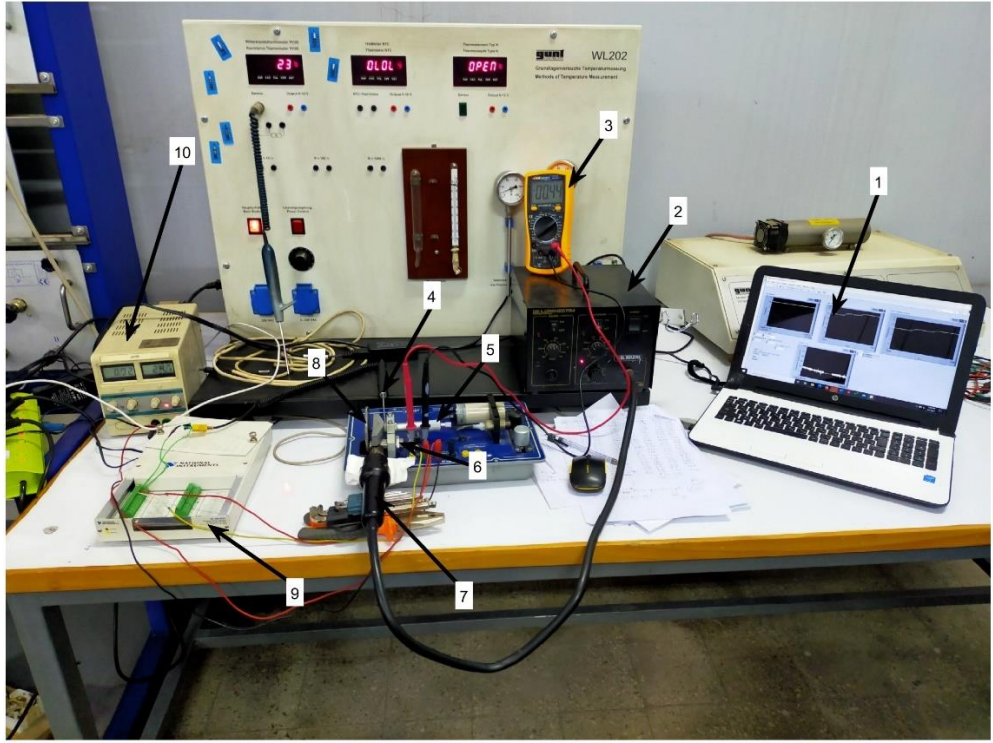

Figure 3. Emulation experiment setup; 1-user interface, 2-heat blower generator, 3-voltmeter, 4-thermocouple, 5-micrometer, 6-inductive sensor, 7-hot air nozzle, 8-steel disk, 9-data acquisition, 10-dc power supply

### 2.2. Machine learning algorithm

The multiple polynomial regression model was developed to capture complex non-linear relationships in the experimental data collected during stage 1. The regression model in this work explicitly represents the dependent variable, clearance (the response) as a function of two independent variables, voltage and temperature. This model consists of a univariate single response, where the clearance serves as the single response variable. The parameter estimates are determined through an iteratively derived closed form equation within the algorithm, ensuring the required performance specifications are met. The experimental dataset, collected multiple times daily in Stage 1 is systematically structured and updated as new measurements are available.

A MATLAB code for a polynomial multiple regression model from the training data is executed. The model algorithm is of the form:

$$d_k = \sum_{i=0}^{m} \sum_{j=0}^{n} a_{ij} V_k^i T_k^j + \epsilon_k, \qquad k = 0,1,\ldots N \tag{1}$$

where $d_k$ is $k^{th}$ data in N training measurement data (measurement of the clearance between the sensor and the wheel flange), $a_{ij}$ represent coefficients of the regression model to be solved, m and n are the orders of the regression model. Clearly, $n$ is the number of features in temperature, and $m$ is the number of features in voltages. V is N-by-1 voltage matrix observations; T is an N-by-1 temperature matrix observations; and $\epsilon$ is an N-by-1 vector of independent identically distributed random measurement errors. The algorithm solves the coefficients by searching coefficients that automatically minimize the cost function which is defined by measurement total squared errors. It means:

$$\frac{\partial}{\partial a_{pq}} \sum_{k=1}^{N} \epsilon_k^2 = 0, \quad or \quad \frac{\partial}{\partial a_{pq}} \left( d_k - \sum_{i=0}^{m} \sum_{j=0}^{n} a_{ij} V_k^i T_k^j \right)^2 = 0 \tag{2}$$

with $p = 0,1,2,\ldots m$ and $q = 0,1,2,\ldots n$

Substituting Equation (1) in Equation (2), performing partial differentiation, and rearranging the terms, Equation (2) becomes:

$$\sum_{k=1}^{N} \sum_{j=0}^{n} \sum_{i=0}^{m} a_{ij} V_k^{i+p} T_k^{j+q} = \sum_{k=1}^{N} d_k V_k^p T_k^q \tag{3}$$

with $p = 0,1,2,\ldots m$ and $q = 0,1,2,\ldots n$

Equation (3) represents the closed form of the dynamic machine learning model which iterates orders (m and n) of the polynomial in features of T and V until a set coefficient of correlation (R-Square) is reached.

Through a set of resulting equations, the coefficients $a_{ij}$ are solved explicitly. The algorithm continuously checks the regression coefficient r of the regression model. Then the conditions for the algorithm performance measure are stipulated as: if r is equal to or greater than 0.975 (a good confidence of 95%), then upload the regression model. Moreover, if the adjusted r is less than 0.975, then increase the order of the polynomial regression model by adding more coefficients till the condition of performance measure is satisfied. Therefore, this algorithm always keeps the minimum order polynomial regression model that satisfies the performance measure conditions. The regression model file stored in the LabView directory is updated every time data is added to the excel file.

The MATLAB script in the LabView program calls this file with the regression equation, and the algorithm executes step-wise instructions. The multiple repeated measurements data ranging from 19.8°C to 73°C are aggregated together and a computer program implemented equation (3) and computed a multiple regression model coefficient.

The example regression model based on our collected input data, which was found to be a second-order polynomial regression model of form,

$$d = 1.373 + 0.2082V - 0.005041T + 0.06928V^2 + 0.002771VT \tag{4}$$

whose coefficients are with good confidence of 97.09%, with adjusted $r_{V,T}^2 = 0.9709$, or $r_{V,T} = 0.9853$. Afterwards, more experimental data with added disk measurements at temperature up to 98°C were added to the database and a second-order regression model with r coefficient of 0.9852 (good confidence of 97.06 %) was found to be:

$$d = 1.16 + 0.3672V - 0.0036T + 0.05812V^2 + 0.0006618VT. \tag{5 a}$$

This algorithm was automatically loaded to the data acquisition software for measurement.

Using a polynomial linear regression, we get:
$$d = -0.3407 + 0.9617V + 0.00595T \tag{5 b}$$
with good confidence of 94.92 %, with adjusted $r_{V,T}^2 = 0.9492$ $r_{V,T} = 0.9743$ was found. The conditions to suffice the 95% good confidence of both models is not satisfied for this case. Here we see that a nonlinear polynomial multivariate regression model performs better that a linear multivariate regression model, which satisfies the suitability of the model in equation (3) to train the system.

## 3. EXPERIMENTAL RESULTS AND DATA ANALYSIS

This subsection discusses the results obtained from the experiments and multi-body simulations. In addition, an online digital filter design with its results is presented.

### 3.1. Effects of temperature on the inductive displacement sensor response

The results presented were obtained from manual measurements recorded on daily basis. The average of these repeated measurements, as illustrated in Figure 4 and Figure 5 depicts the relationship between sensor output and the clearance under measurement, for various environment temperatures. The challenge while taking these measurements was keeping the temperature constant while collecting data. The drift of the measurement due to temperature changes is seen. From the room temperature, as temperatures around the inductive sensor are increased starting from room temperature (20°C), the voltage output decreases. Increasing temperatures from 60°C to 73°C, there is a minimal drift from measurement data in span from 0 mm to 5 mm. For measurements ranging is registered, and from 5 mm to 10 mm, a significant temperature effect on data is observed. It is seen that as the voltage diverges.

Further, increasing temperature from 79°C to 98°C results in a decrease in voltages corresponding to measurements ranging from 5 mm to 10 mm (see Figure 5). There is consistent drop in the analog output voltage as the temperature around the inductive sensor is increases. This phenomenon can be explained by the nonlinear relationship between temperature variation and the impedance of the ferrite coil in the shielded analog inductive sensor decreasing. As the temperature rises, the impedance decreases, leading to a reduction in the sensor's output voltage.

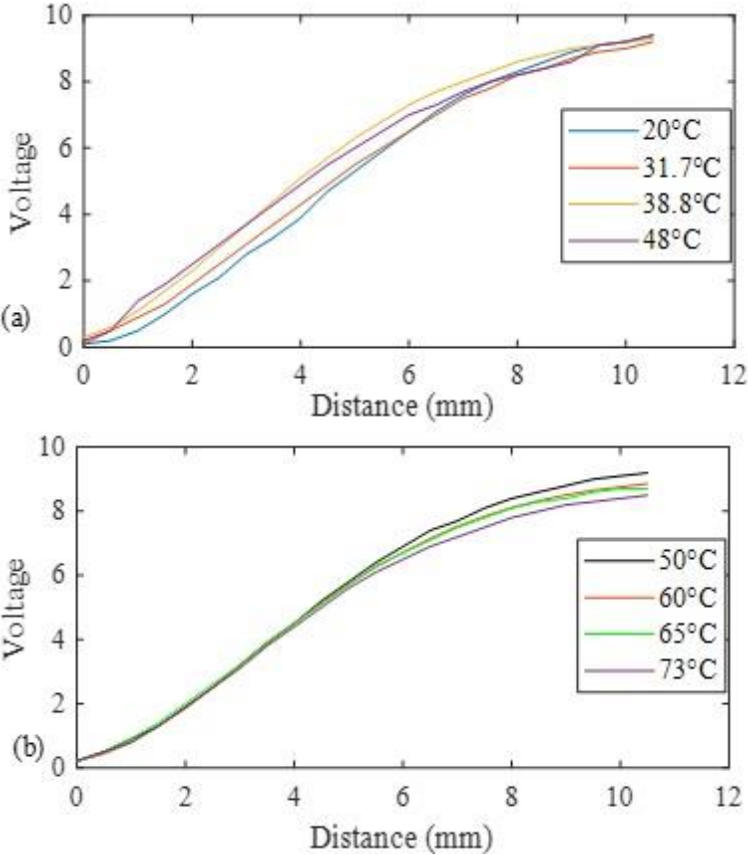

Figure 4. Average daily measurements from $19.8°C$ $to$ $73°C$

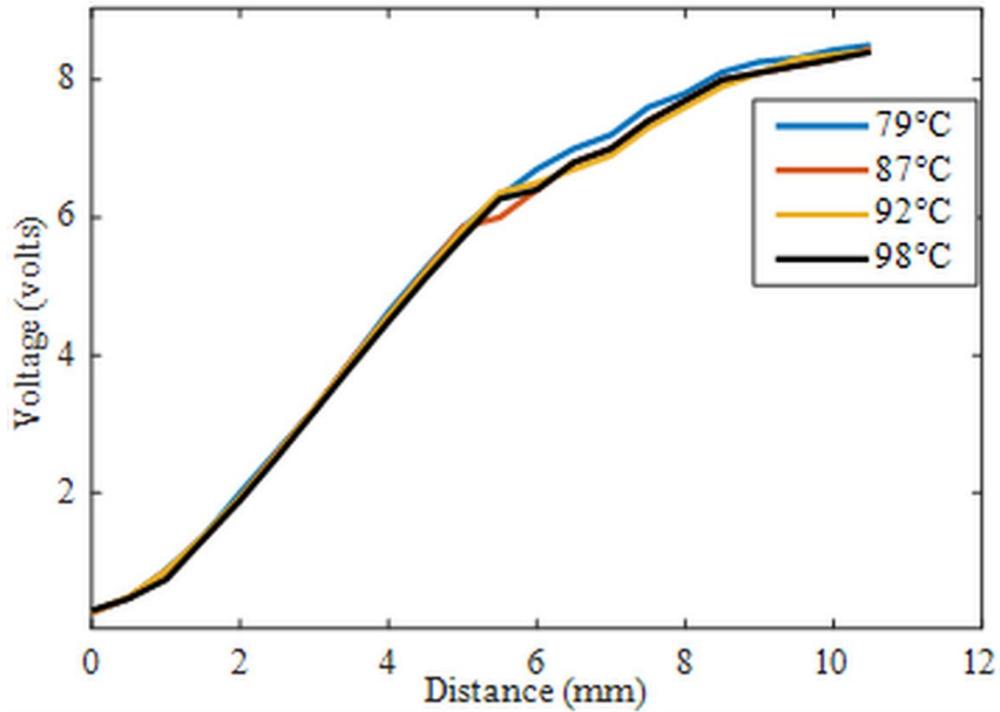

Figure 5. Average daily measurements from 79°C to 98°C

### 3.2. Error in measurement relative to clearance at different temperatures

This subsection analyzes measurement errors in the output voltage of the inductive displacement sensor at different temperatures. The errors are evaluated for each clearance measurement, with the reference measurement taken at room temperature, where the sensor operates under standard conditions. During laboratory experiments, the sensor's normal operating temperature was $20°C$. The error in output voltage at each temperature is determined using the following formula.

$$error\ in\ measurement = |V - V_R|, \quad (6)$$

where $V$ is the observed output voltage at each clearance and $V_R$ is the reference voltage that is taken at room temperature of 20°C. Figure 6 presents a comparison of errors in measurement from 20°C to 65°C. The error at 20°C is at the baseline zero in the graphs since it is taken to be the reference point. From the presented results, we see the variation in errors in measurements at different temperatures.

Further, we continue to determine the error in measurement at elevated temperatures, that is, from 20°C to 98°C (see Figure 6 and Figure 7). It is evident that errors in measurement due to the temperature effect are significant, hence this calls for machine learning methods to train the system on the patterns of the measurements with different temperature input conditions.

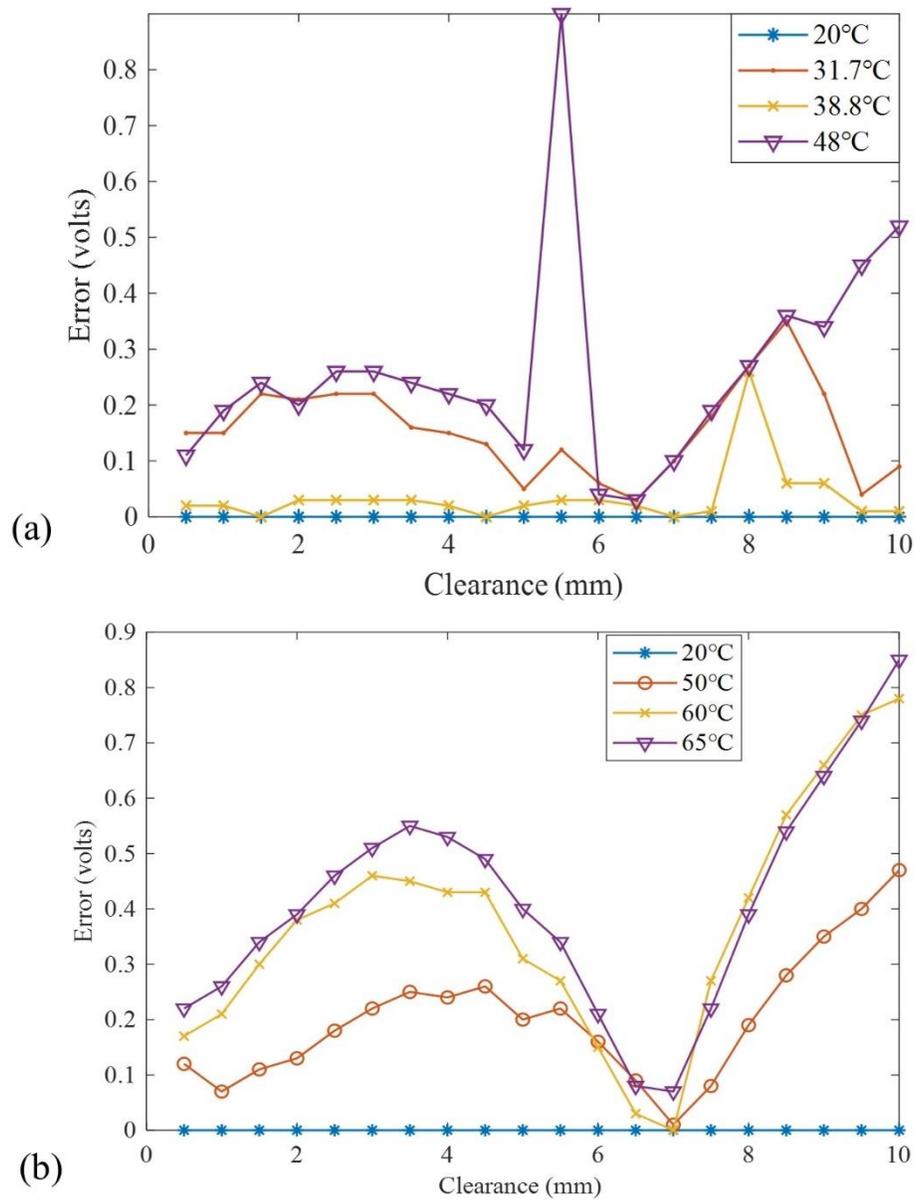

Figure 6. Error in measurement at different temperature

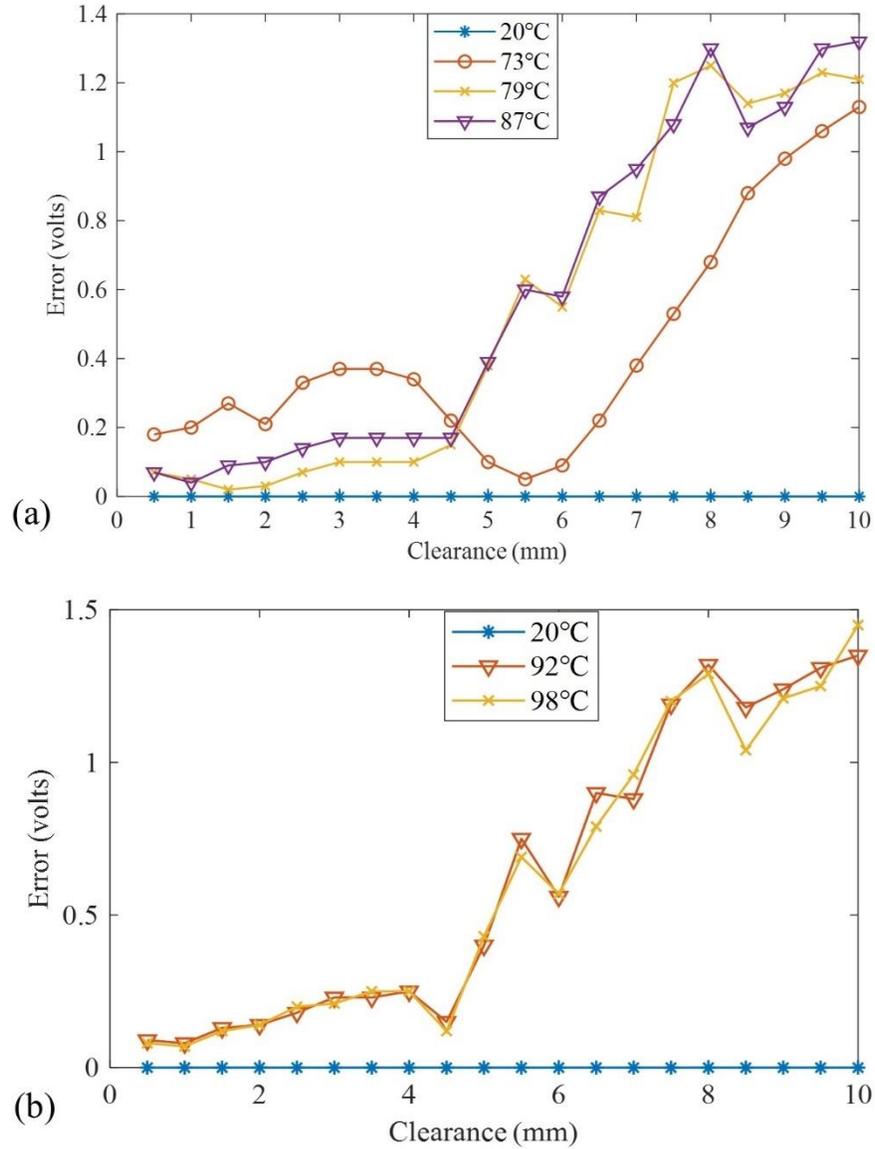

Figure 7. Error in measurement relative at different temperatures

### *3.3.* **Response of the measurement system**

Stage two of the experiments, involving the data acquisition system presents measurement results of predicted clearance over time. The system's measurement response is obtained from the fused data of inductive sensor and type K thermocouple, using a regression equation that is regularly updated as new data are uploaded to the database. Figure 8 shows the measurement system response taken at temperatures ranging from 43℃ to 50℃ ("Measured Values"). Simultaneously, manual measurements ("True values") are taken using a micrometer and a voltmeter at predefined chronometer time intervals. By comparing both measurements, the accuracy of the measurement system is determined.

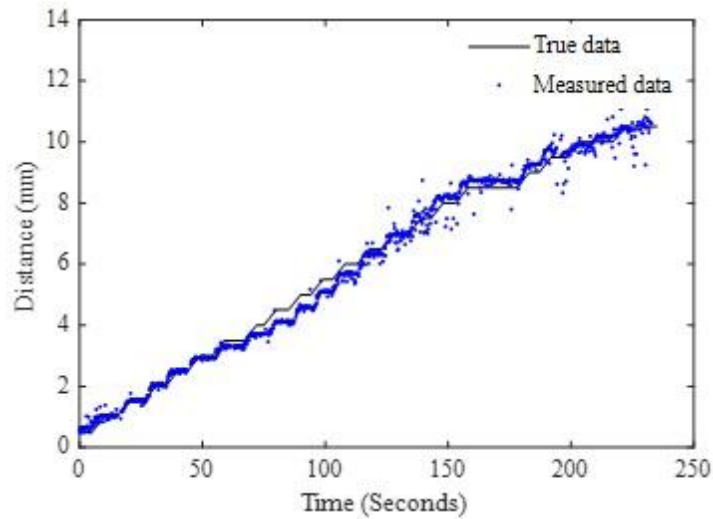

Figure 8. Measured distance readings against time using the data acquisition system. The recorded distance represents the clearance between the sensor tip and the wheel flange's rail gauge-facing surface.

Accuracy of the system is calculated in terms of absolute error and is provided in Table 1:

$$\text{percent error} = [(|d_o - d_a|)/d_a]\%, \tag{7}$$

where, $d_o$ = Measured value, $d_a$ = True value,

$$|d_o - d_a| = \text{Absolute error}$$

Table 1. Accuracy estimation of the system at different disk position

| Clearance (mm) | Error | %error |
|---|---|---|
| 1 | 0.02595 | 2.595 |
| 2 | 0.020736 | 2.0736 |
| 3 | 0.024471 | 2.4471 |
| 4 | 0.078376 | 7.8376 |
| 5 | 0.081988 | 8.1988 |
| 6 | 0.056695 | 5.6695 |
| 7 | 0.01452 | 1.452 |
| 8 | 0.02922 | 2.922 |
| 9 | 0.027095 | 2.7095 |
| 10 | 0.0226914 | 2.26914 |

The system achieved an average accuracy of 96.1826%. Additional measurements recorded over temperature range of 19.8°C to 98°C were added to the database, and the regression model was subsequently updated (See Equation 5). The system then took measurements over temperatures range of 92°C to 96°C, and the predicted (emulated) clearance is as shown in Figure 9. The accuracy of the system was further evaluated by comparing the predicted clearance by the dynamic machine learning algorithm against manual measurements taken using a stop watch and micrometer as the disk was displaced.

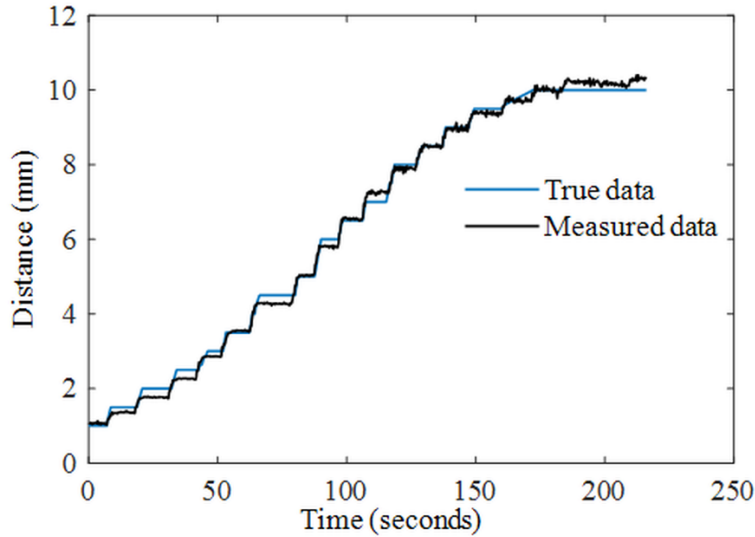

Figure 9. Accuracy test of the acquitted measurement

Equation (8) determines the system's accuracy at each disk position in terms of percentage error. Table 2 Presents the results, which are averaged to get the overall system accuracy at elevated temperatures.

Table 2. Accuracy estimation of the system at different clearances

| CLEARANCE (mm) | error | Percent error (%) |
|---|---|---|
| 1 | 0.068014 | 6.801436 |
| 1.5 | 0.089082 | 8.908169 |
| 2 | 0.107114 | 10.71144 |
| 2.5 | 0.084953 | 8.495297 |
| 3 | 0.044563 | 4.456256 |
| 3.5 | 0.014637 | 1.463719 |
| 4 | 0.025938 | 2.593796 |
| 4.5 | 0.047186 | 4.71855 |
| 5 | 0.007226 | 0.722596 |
| 6 | 0.031677 | 3.167742 |
| 6.5 | 0.007587 | 0.75868 |
| 7 | 0.034893 | 3.489255 |
| 7.5 | 0.009612 | 0.961197 |
| 8 | 0.012693 | 1.269339 |
| 8.5 | 0.00301 | 0.301005 |
| 9 | 0.006372 | 0.637185 |
| 9.5 | 0.012563 | 1.256268 |
| 10 | 0.016844 | 1.684358 |
| **Average** | 0.034665 | 3.46646 |

The system's accuracy from the predicted response at different disk positions with the added data has improved to 96.5%.

## 4. ONLINE FILTER DESIGN AND SIMULATION

As observed in Figure 8, there are noises that can be present in measurement resulting from systematic errors during measurements. These noises are represented by higher frequencies observed in the Fast Fourier Transform (FFT) results of the sensor output signal (clearance). The frequency domain signal of the clearance is correlated with the actual clearance measurement in frequency domain to determine the filter design specifications (Figure 10). The actual clearance measurement is obtained by using hand tools (micrometer readings at chronometer time intervals as shown in Figure 9).

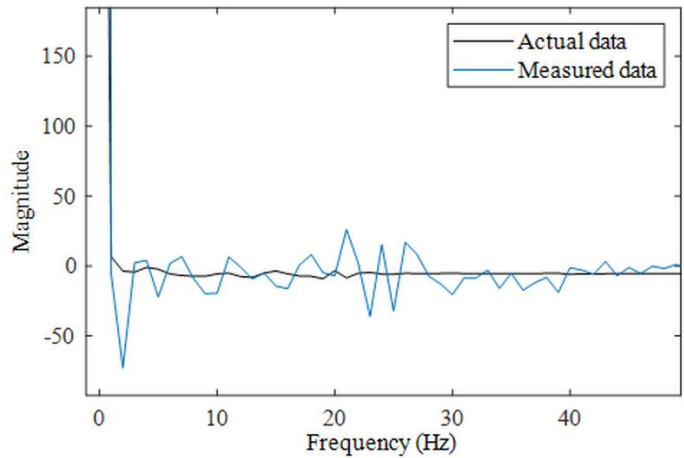

Figure 10. FFT analysis of the real-time measurements and actual true measurements

This measurement filtering system is chosen to be a recursive, whose an output at any instant depends on a set of values of the input as well as a set of values of the output. The response of a fairly general recursive, linear time invariant, discrete-time filter is given by (9)

$$y(n\tau_s) = \sum_{i=-K}^{M} a_i x(n\tau_s - i\tau_s) - \sum_{i=1}^{N} b_i y(n\tau_s - i\tau_s), \qquad (8)$$

where $\tau_s$ is the sampling period, n is the number of samples at a given time, y is the filter output signal, x is the filter input signal. That is, if at an instant $nT$ were taken to be present, then the present response would be a function of the past M values, the present values, the future K values of the excitation as well as the past N values of the response. The dependence of the response on a number of past values of the response implies that a recursive discrete-time system must involve feedback from the output to the input. A low pass Butterworth recursive filter, is selected to be designed to remove out the noises from the measurement data. Its suitable characteristic is one of maximum flatness in the passband at frequency zero, and maximum flatness in the stopband at infinite frequency. The transition band roll-off rate, however, is very shallow for a given filter order N. However, in practice Butterworth ideal frequency response is unattainable as it produces excessive passband ripple (30) . The FFT analysis specifies the filter design specification.

A generalized equation of the frequency response amplitude representing a nth order Butterworth filter is defined by:

$$|H_n(j\omega)|^2 = \frac{1}{1 + \left(\frac{\omega}{\omega_c}\right)^{2N}}, \qquad (9)$$

where $\omega_c$ is the upper cut-off frequency in radians per second, $\omega$ is the frequency of the input signal, and N is the order of the filter. The squared response of the Butterworth filter is given as the function of the cut-off frequency. The design specifications from FFT analysis of the output signal are given as: sampling frequency $f_s$= 6 Hz or sampling period as $\tau_s = 0.1667$ sec. Approximation of the passband edge $f_p = 0.1\ Hz$, and stopband edge $f_a = 0.9\ Hz$, are obtained from the frequency domain clearance output signal to be filtered (see Figure13). We approximate the cut-off frequency (rad/sec) as $f_c = f_a$.

Next, we apply a pre-warping transformation, pre-warp critical (stopband, passband) frequencies to transform to a required continuous system, for which $\omega$ is in the digital domain and $\Omega$ is in the analog domain.

Therefore, the pre-warped critical frequencies are:

$$\Omega_p = \frac{2}{\tau_s} \tan \frac{\omega_p \tau_s}{2} \tag{10}$$

$$\Omega_a = \frac{2}{\tau_s} \tan \frac{\omega_a \tau_s}{2} \tag{11}$$

In which, $f_p = 0.21128 Hz$ and $f_a = 0.975 Hz$ shows the warping effect near $\frac{f_s}{2}$.

### 4.1. Minimum order filter design

For efficient operation, without response delays of the measurement system, the filter must respond faster. Lower is the order of the filter faster it responds, but the less the filter signal output follows the input in the passband, and the less attenuation it achieves in the stopband. The higher the order of the filter is, the higher the filter output follows the filter input and the higher attenuation it achieves in the stopband. Here we design the minimum order filter that achieves the prescribed specifications that are the maximum passband attenuation and the minimum stopband attenuation. From the frequency spectrum of measured clearance before applying filtration, we chose maximum passband attenuation of $A_p = 0.1dB$, and minimum passband attenuation of $A_a = 40dB$

The minimum order of the Butterworth filter that achieves aforementioned specifications is given by:

$$N = \frac{\log \frac{\sqrt{10^{0.1 A_a} - 1}}{\sqrt{10^{0.1 A_p} - 1}}}{\log \left( \frac{\Omega_a}{\Omega_p} \right)}, \tag{12}$$

$$\Omega_c = \frac{\Omega_p}{\left( \sqrt{10^{0.1 A_p} - 1} \right)^{\frac{1}{N}}}. \tag{13}$$

Table 3 indicates the numerical values of the filter specification quantities that lead to determining the filter order and cut-off frequency.

Table 3. Filter specification parameters

| Symbol of the variable | Variable name | Calculated values | unit |
|---|---|---|---|
| $\omega_a$ | Continuous signal stop band edge frequency | $1.8\pi$ | Rad/sec |
| $\omega_p$ | Continuous signal passband edge frequency | $0.2\pi$ | Rad/sec |
| $\omega_c$ | Continuous signal cut-off frequency | 3.142 | Rad/sec |
| $A_a$ | Minimum stopband attenuation | 40 | dB |
| $A_p$ | Maximum Passband attenuation | 0.17 | dB |
| $\Omega_p$ | Pre-warped passband edge frequency | 1.328 | rad/sec |
| $\Omega_a$ | Pre-warped stopband edge frequency | 6.1287 | rad/sec |
| $N$ | The filter minimum order | 5 | integer |
| $\Omega_c$ | Pre-warped cut-off frequency | 1.871 | rad/sec |

Upon determining the order of the filter and the cut-off frequency, the next step is to determine the poles of the filter transfer function. The filter transfer function amplitude shown in equation (9) is obtained from the transfer function:

$$H(jP) = \frac{1}{1+(jP)^N} \quad \text{or} \quad H_n(s) = \frac{1}{C(s)} \qquad (14)$$

where $s = jP = j\frac{\omega}{\omega_c}$ and C(s) is the characteristic equation of the filter transfer function. The characteristic equation is obtained by a product of binomial terms $s - s_i$ for i=1, 2, 3, 4… N, where $s_i$'s are the poles of $\frac{1}{1+(jP)^N}$ in $jp$ variable. The predetermined transfer function of a normalized fifth order low pass filter is given as:

$$H_5(s) = \frac{1}{s^5 + 3.236s^4 + 5.236s^3 + 5.236s^2 + 3.236s + 1}, \qquad (15)$$

Now we substitute $s \to \frac{s}{\Omega_c}$ to get the filter transfer function of the specified cut off frequency:

$$H_n\left(\frac{s}{\Omega_c}\right) = \frac{1}{0.0014s^5 + 0.0168s^4 + 0.1011s^3 + 0.377s^2 + 0.8683s + 1}, \qquad (16)$$

Applying the bilinear transformation to the transfer function in equation (15) of an analog filter yields a digital filter characterized by the discrete-time transfer function:

$$H(z) = H(s)\Big|_{s=\frac{2}{\tau_s}\left(\frac{1-z^{-1}}{1+z^{-1}}\right)}, \qquad (17)$$

where $\tau_s = 0.167$ seconds. Finally, we get the digital filter of the following transfer function in z domain:

$$H(z) = \frac{0.001076 + 0.005379z^{-1} + 0.01076z^{-2} + 0.01076z^{-3} + 0.005379z^{-4} + 0.001076z^{-5}}{1 - 3.06z^{-1} + 3.997z^{-2} - 2.721z^{-3} + 0.9566z^{-4} - 0.138z^{-5}} \quad (18)$$

Figure 11 shows the magnitude response of the fifth-order filter

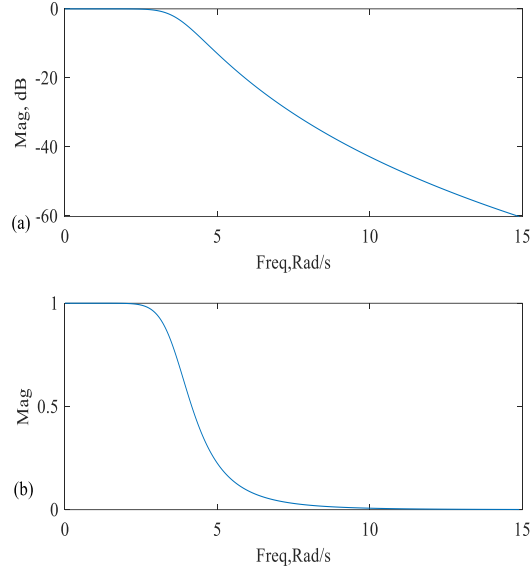

Figure 11. (a) Magnitude (dB) response, and (b) Magnitude (linear) response

It is known that transfer function;

$$H(z) = \frac{Y(z)}{X(z)}, \text{ from which } Y(z) = H(z)X(z)$$

Inversing the transfer function, the finite difference equation of the filter is found to be:

$$y(n) = 3.06y(n-1) - 3.997y(n-2) + 2.721y(n-3) - 0.9566y(n-4) + 0.138y(n-5) + 0.001076x(n) + 0.005379x(n-1) + 0.01076x(n-2) + 0.010706x(n-3) + 0.005379x(n-4) + 0.001076x(n-5) \quad (19)$$

### 4.2. Filter design and simulation results

From the filtered data, we compute the accuracy of the signal to the actual disk displacements. The accuracy of the filtered data with reference data has improved, as shown in Table 4.

Table 4. Calculated accuracy of the filtered data

| Distance (mm) | Error | %error |
|---|---|---|
| 1 | 0.015534 | 1.5534 |
| 2 | 0.019077 | 1.9077 |
| 3 | 0.021802 | 2.1802 |
| 4 | 0.070238 | 7.0238 |
| 5 | 0.078392 | 7.8392 |
| 6 | 0.052738 | 5.2738 |
| 7 | 0.004391 | 0.4391 |
| 8 | 0.01114 | 1.114 |
| 9 | 0.027036 | 2.7036 |
| 10 | 0.011554 | 1.1554 |

The actual data with filtered online data are compared, from which absolute to measurement error of the system was averaged to 97.2%.

## 5. MULTI-BODY SYSTEM (MBS) SIMULATION AND FINAL FILTER DESIGN

### 5.1. Locomotive multi-body system (MBS) dynamics simulation

When the train moves on track, many dynamics involve changing the wheelsets and the bogie frame positions. With the sensor support structure fixed on the bogie frame and in face of the wheel flange, these dynamics affect measurement readings as the set point of the inductive sensor may be changing. The previous research (13) studied lateral and yaw displacements of the wheelsets at curves about the designed sensor holding fixture. Here we look at vertical, lateral, and yaw displacements of the wheelsets and bogie frame. The Ethio-Djibouti railway under study uses a co-co locomotive to drive both the passenger and freight wagons. This co-co locomotive has two bogie frames, each of which has three wheelsets. The vertical, lateral, and yaw displacements of the bogies and wheelsets translates noises in the flange measurements through sensor support structure fixed on the bogie frame. The multi-body simulations are carried out to determine these displacements when the locomotive moves through curved tracks and show vibrations translated from the irregularities of the track. Therefore, a suitable sensor support mechanism design is proposed to maintain the set coordinate of the sensing tip during curves. An online digital filter to removing noises in measurement resulting from displacements, locomotive vibrations, and track irregularities is designed. Equations governing the locomotive dynamics are defined about the existing principles (31). D'Alembert's principle is used to define the equations of motion of the Locomotive by using the system coordinates moving along the curved track with vehicle speed. The locomotive equation of motion is described as a second-order nonlinear differential matrix equation in the time domain.

$$m_v a_v + c_v(v_v)v_v + k_v(x_v)x_v = F_v(x_v, v_v, x_t, v_t) + F_{ext} \qquad (20)$$

where $x_v, v_v$, and $a_v$ are vectors of displacements, velocities, and accelerations of the rail vehicle, respectively; $m_v$ is the mass matrix of the vehicle; $c_v(v_v)$ and $k_v(x_v)$ are the damping and stiffness matrices which depend on current state of the rail vehicle, hence describing nonlinearities within the suspension; $x_t$ and $v_t$ are the vectors of displacement, and velocities of the track subsystem; $F_v(x_v, v_v, x_t, v_t)$ is the system load vector representing the non-linear wheel-rail contact forces that depend on the motion displacements and velocities, $x_v$ and $v_v$ of the rail vehicle and $x_t, v_t$ of the track; and $F_{ext}$ are external forces including gravitational forces and centripetal forces when the Locomotive moves through a curve. The computation of wheel rail interaction forces is based on Kalker "Variational and Numerical Theory of Contact" (32). It assumes small creepages for the computation of creep forces. This assumption ensures that the creep forces remain within the linear range (33,34), providing reliable and simplified approach to modelling wheel-rail interactions. It is particularly applicable to scenarios where wheel flange wear remains within mild or moderate ranges (34). In the locomotive model, there are several principal components, including one loco-body, two bogies, six wheelsets, traction rods, reduction gearboxes, rotor Cardan shaft, and axle boxes. All components are assumed to be rigid each with 5 degrees of freedom (vertical, lateral, yaw, pitch and roll motions). With all the components inclusive, the Locomotive is modeled 106 degrees of freedom. Figures 12 and 13 show the locomotive model and co-co bogie designed using the universal mechanism software.

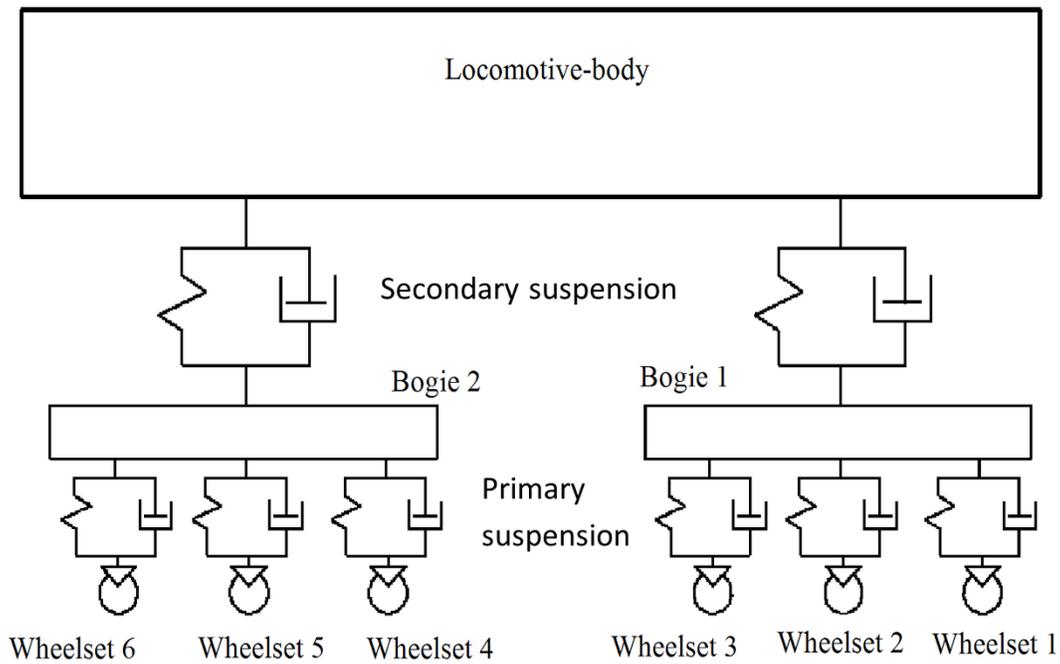

Figure 12. Two-dimensional lamped-parameter locomotive model: side elevation

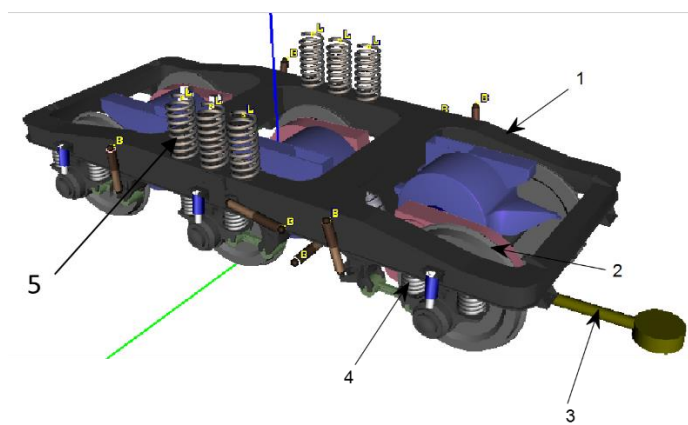

Figure 13. Bogie model: bogie frame (1), wheelset with motoring assembly (2), traction rod (3), primary suspension (4), secondary suspension (5)

The following parameters are considered for the track model:

- A tangent section of 10 m.
- The length of the spiral (transition curve between tangent track and curved track) is set to 50 m before the curve of the constant radius.
- The curved track section of constant radius has a length of 300 m.
- The cant (track superelevation) is set to 0.09 m.
- The spiral transition curve after the curved track, connecting back to the tangent track, is also set to 50 m.
- Track widening in a curve is specified as: 15 mm for $R < 300\ m$, and $10\ mm$ for $R \leq 350\ m$.

Additionally, the point (0;0;1.4) is selected on two bogie frames, and for each bogie, one wheelset is selected with the sensor pointed at wheelbase (0;0;0).

We will run the locomotive at a velocity of 20 m/s on track with the single lateral irregularity at the beginning of the track. The amplitude of the irregularity is 20 mm, and its length is 10 m. We will observe the effects of these irregularities from the displacement results presented. Specifying the track properties as stated, a graphical track model is shown below (Figure 14).

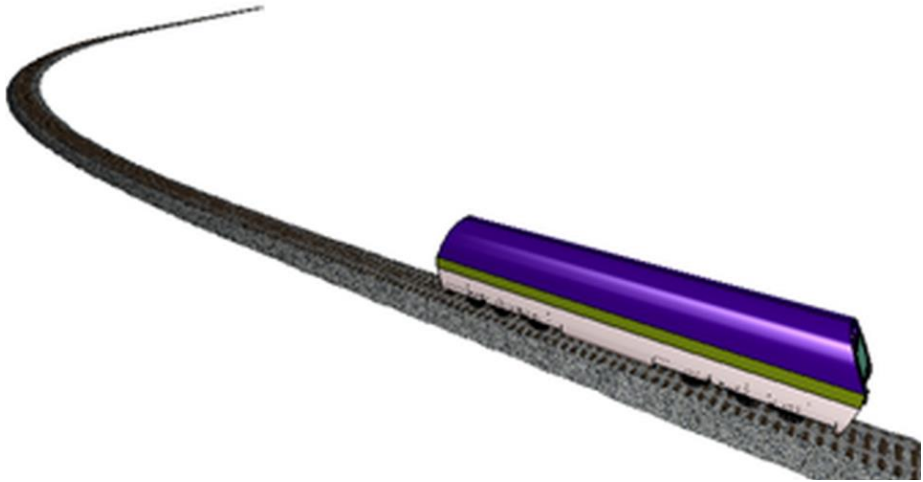

Figure 14. Locomotive moving through the curved track

It is worth noting that the testing of the measurement system so far done did not include dynamic effect of the rail vehicle. In [11], only a single wheelset dynamic was simulated to quantify the effect the sensor readings. In this work, we consider the full effect caused by the locomotive dynamic on sensor readings. It is noted that the sensor is rigidly fixed on the truck structure of the locomotive. It is emphasized that the relative vertical displacement between the wheelset and the locomotive truck does not have effect on the clearance between the sensor and the wheel flange. It is also noted that there is no relative lateral motion between the wheelset and the locomotive truck. However, there is a relative yaw motion between the bogie and the wheelset. This relative yaw motion will have an effect on the clearance between the sensor and the wheel flange. If this effect is predetermined during train operation, it can be removed from the measurement data by further filtering. This subsection presents the multi-body simulation results of the vertical, lateral, and Yaw displacement motions of the wheelset and the bogie. Frequency analysis of these displacements is done to where they are correlated with the measurement data.

### 5.2. Vertical, lateral, and yaw displacements

All the parameters to be analyzed are selected in the simulations of the MBS locomotive model. Below are the graphical results data showing vertical, lateralis the graphical results data showing vertical, lateral, and yaw motion of wheelsets and bogie. Vertical displacement of the left and right wheels on all wheelsets of leading bogie one are represented in Figure 15. During the curve transition, oscillations due to the lateral irregularity with amplitude 20 mm at length 10 m are shown by the uniform trend of the curve. At constant curve radius, vertical displacement increases from the initial position, where the point of wheel-rail contact is displaced, thus the wheel is displaced as the train moves through the curve. When the Locomotive nears the end of transition curve towards the straight track the positions of the wheelsets come to initial positions.

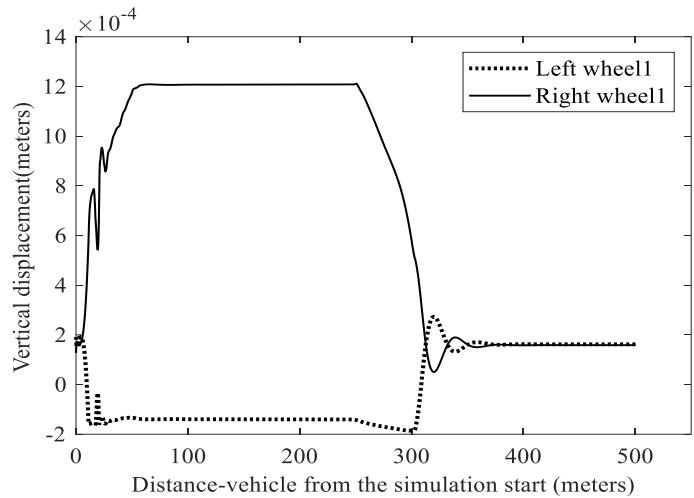

(a)

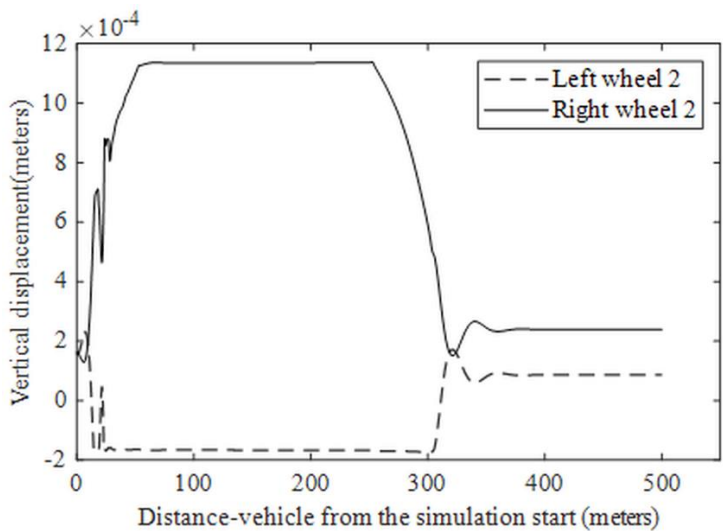

(b)

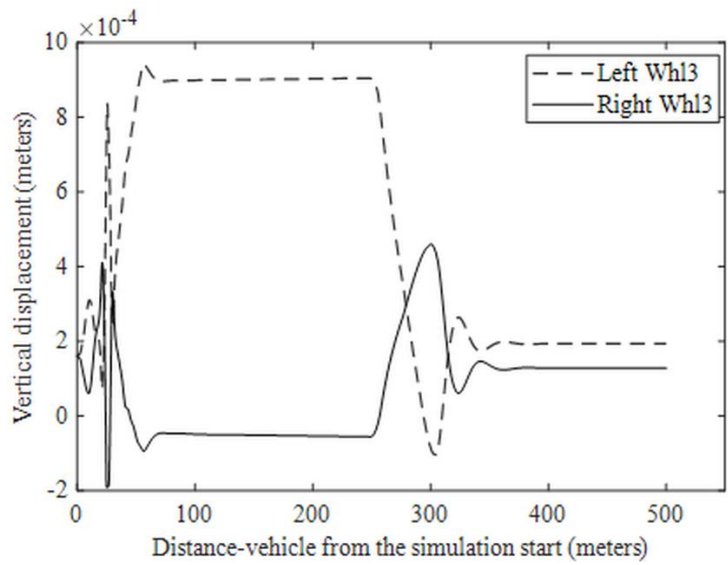

(c)

Figure 15. Vertical displacement of the wheelsets (Whl3: wheel set 3): (a) Wheel set 1, (b) wheel set 2, (c) wheelset 3

Table 5 presents maximum vertical displacements of the left and right wheels for three wheelsets

Table 5. Maximum vertical wheelset displacement

| Wheelset number | Vertical Displacement | |
|---|---|---|
| | Right Wheel (mm) | Left wheel (mm) |
| 1 | 1.2 | 0.2 |
| 2 | 1.135 | 0.1672 |
| 3 | 0.188 | 0.9322 |

With the primary suspensions, the bogie vertical displacement from its sitting position increases to 34 mm on the constant curve radius (see Figure 16).

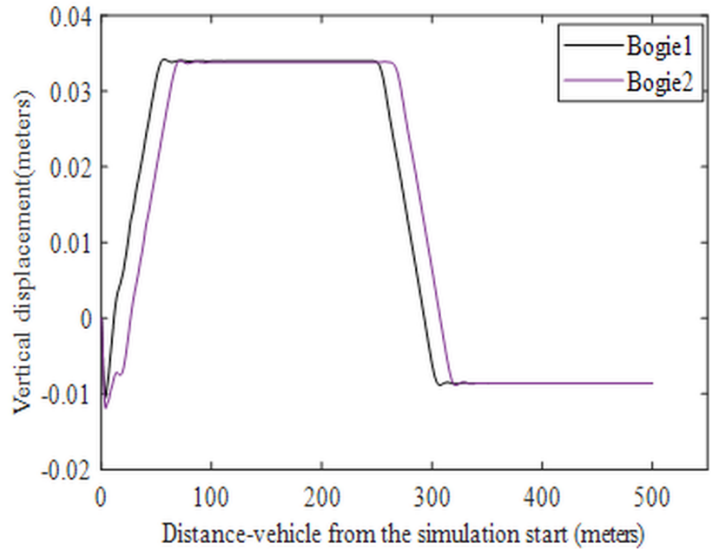
Figure 16. Vertical displacement of the bogie

The centripetal forces experienced at the curve shifts the positions of wheels at different points from initial contacts. The three wheelsets have different lateral displacement values. Wheelset one and two seems to have the same motion different from the three (see

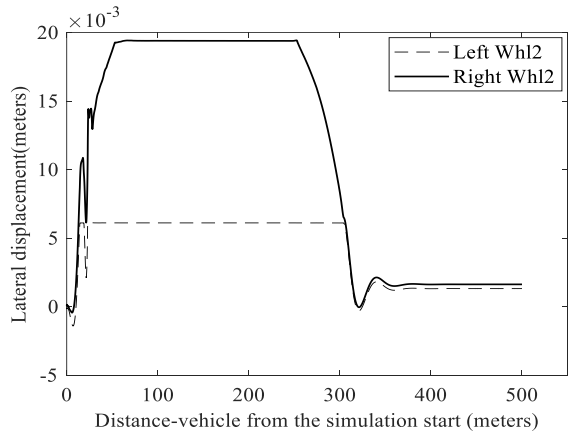

(b)

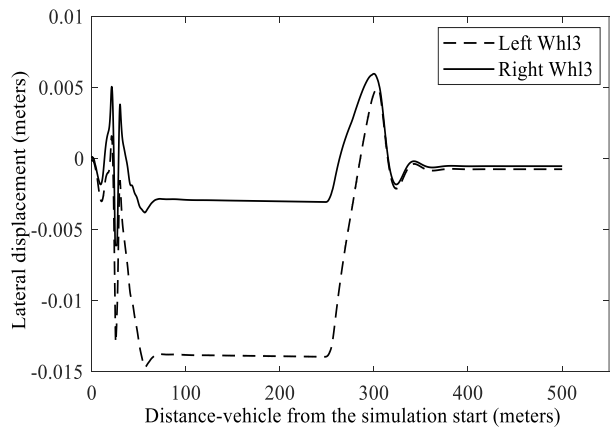

(c)

Figure 17). Table 6 shows the maximum lateral displacements of both left and right wheels.

Table 6. Maximum lateral wheelset displacement

| Wheelset | Lateral Displacement | |
|---|---|---|
| | Right Wheel (mm) | Left wheel (mm) |
| 1 | 21.41 | 6 |
| 2 | 19.4 | 6 |
| 3 | 5.92 | 14.64 |

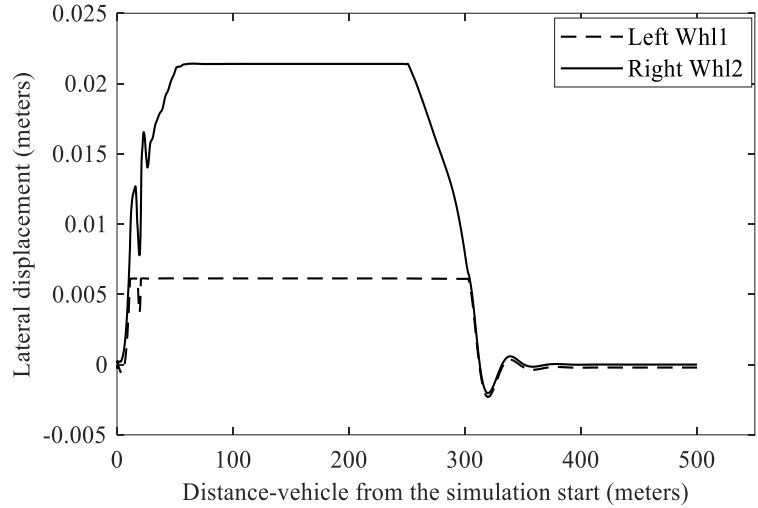

(a)

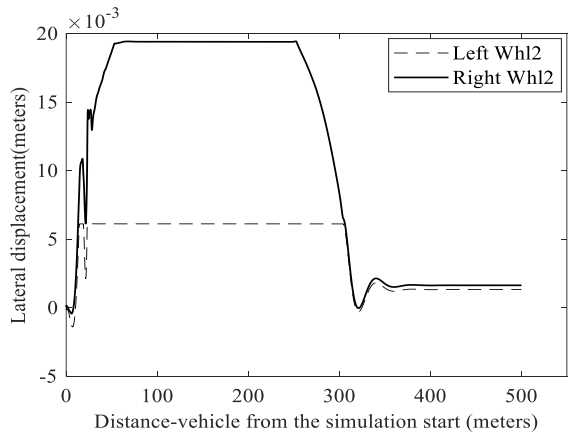

(b)

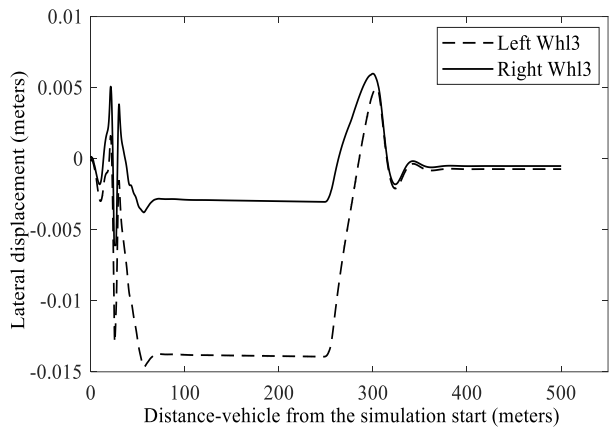

(c)

Figure 17. Lateral displacement of the wheelsets. (a) First wheelset, (b) second wheelset, (c) third wheelset

The yaw motion of the wheelsets are presented in Figure 18. Vibrations from the lateral irregularity are seen at the beginning when the train is in a transition curve. Wheelset one leads in the yaw displacements on the two bogies.

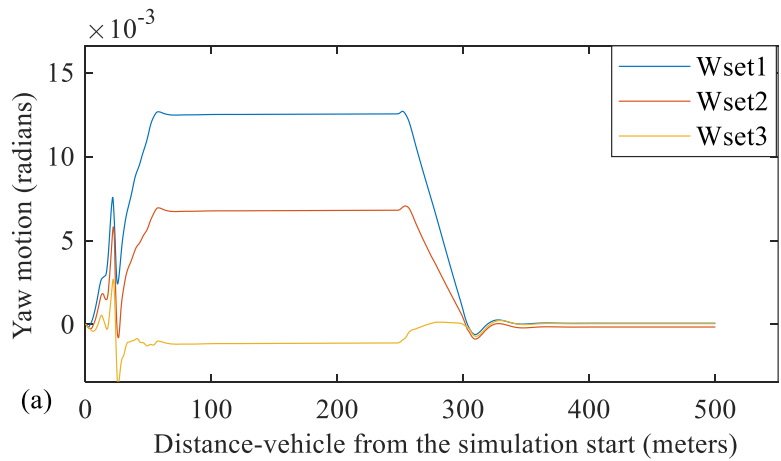

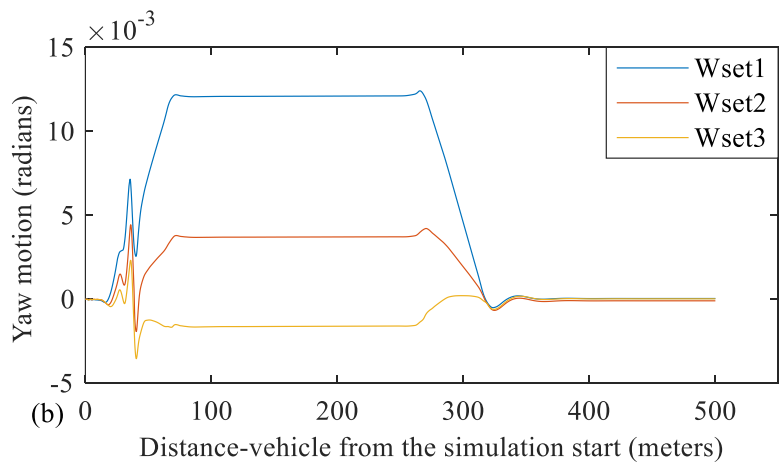

Figure 18. Yaw motion of the wheelsets. (a) front bogie, (b) rear bogie

Table 7 tabulates the maximum yaw displacement of the wheelsets relative to the local system of coordinate at the bogies

Table 7. Yaw displacement

|   | Maximum Yaw displacement (radians) | |
|---|---|---|
|   | Bogie 1 wheelsets | Bogie 2 Wheelsets |
| 1 | 0.012578 | 0.012 |
| 2 | 0.00608 | 0.0037137 |
| 3 | 0.00116 | 0.001577 |

### 5.3. Frequency analysis of the displacements

The FFT analysis of the displacements is correlated with actual measurements. Higher frequencies resulting from vibrations from the locomotive dynamics that will affect measurement data are identified, and stopband attenuation in decibels is determined. A stopband frequency of 50 Hz corresponding to the stopband amplitude ripples ($\delta$) is chosen for all the FFT analysis of displacements.

Figures A1, A2, and A3 (see appendix) represent the FFT analysis of displacements from which the filter specification are designed. Table 8 shows the stopband attenuations of the wheelsets and bogie displacement frequencies.

Table 8. Stopband attenuation for the displacement frequencies

| | Stopband Attenuation (dB) | | |
|---|---|---|---|
| | Vertical | Lateral | Yaw |
| Wheelset1 | 47.69 | 27.81 | 20 |
| Wheelset2 | 60 | 24 | 21 |
| Wheelset3 | 60 | 30.75 | 21.22 |
| Wheelset4 | 52.69 | 30.18 | 23 |
| Wheelset5 | 53.97 | 20.94 | 21.87 |
| Wheelset6 | 49.77 | 17.78 | 23.5 |
| Bogie1 | 23.769 | 35.65 | |
| Bogie2 | 23.769 | 35.65 | |

The vertical displacement amplitude is smaller than that of the lateral and Yaw motion, meaning there are more vibrations in the lateral and Yaw motions. $\delta$ of the vertical wheelsets range from 0.001 to 0.004, while for the bogie frame where the sensor support structure is fixed equals 0.0648. Lateral displacement of the wheelsets registers much higher amplitudes than the vertical motions. The bogie lateral movement amplitudes are slightly lower than those of vertical motions. $\delta$ of the lateral wheelsets range from 0.04 to 0.129, while for the bogie frame equals 0.0165. Yaw motions frequency amplitudes shoot from 0.1 to 0.0668. Therefore, to remove these higher frequencies that affect the measurement, a low pass digital Butterworth filter with standard specifications to remove all these unwanted frequencies from all the vehicle dynamics. Lower amplitudes of the stopband attenuations with lesser ripples are chosen to design this filter.

### 5.4. Final Filter Design

The data storage rate of the multi-body simulation was 0.01seconds for every displacement analysis; hence, sampling frequency $(f_s)$= 100 Hz. Passband edge $(f_p) = 10\ Hz$ and stopband edge $(f_a) = 50\ Hz$ are selected from the FFT analysis. Maximum passband loss $A_p = 0.02\ dB$, minimum stopband loss $A_a = 60\ dB$. Equations (11)-(15) are used to calculate the digital cut-off frequency and the minimum filter order. All the filter specification parameters are tabulated in Table 9. A filter order of 7 was attained. As in equations (15)-(19), the same procedure used to attain equations (21)-25).

Table 9. Filter specifications

| Symbol of the variable | Variable name | Calculated values | unit |
|---|---|---|---|
| $\omega_a$ | Continuous signal stop band edge frequency | 314.159 | rad/sec |
| $\omega_p$ | Continuous signal passband edge frequency | 62.832 | rad/sec |
| $\omega_c$ | Continuous signal cut-off frequency | 188.495 | rad/sec |
| $A_a$ | Minimum stopband attenuation | 0.02 | dB |
| $A_p$ | Maximum Passband attenuation | 60 | dB |
| $\Omega_a$ | Pre-warped stopband edge frequency | 1.1 | rad/sec |
| $\Omega_p$ | Pre-warped passband edge frequency | 5.48 | rad/sec |
| $N$ | The filter minimum order | 7 | integer |
| $\Omega_c$ | Pre-warped cut-off frequency | 1.626 | rad/sec |

The predetermined transfer function of a normalized seventh-order low pass filter is given as:

$$H_n(s) = \frac{1}{s^7 + 4.494s^6 + 10.0978s^5 + 14.5918s^4 + 14.5918s^3 + 10.0978s^2 + 4.494s + 1} \quad (21)$$

Substituting for $s \to \frac{s}{\Omega_c}$ in equation (21), where $\Omega_c$ is calculated using equation (13), we get:

$$H(s) = H_n\left(\frac{s}{\Omega_c}\right),$$

$$H(s) = \frac{1}{0.03314s^7 + 0.2423s^6 + 0.8857s^5 + 2.083s^4 + 3.388s^3 + 3.815s^2 + 2.762s + 1} \quad (22)$$

Applying the bilinear transformation to the transfer function of an arbitrary analog filter yields a digital filter characterized by the discrete-time transfer function

$$H(z) = H(s)\big|_{s=\frac{2}{T}\left(\frac{1-z^{-1}}{1+z^{-1}}\right)},$$

$$H(z) = \frac{\sum_{i=0}^{7} a_i z^{-i}}{\sum_{i=0}^{7} b_i z^{-i}} \quad (23)$$

### 5.5. Final filter design results

The coefficients of the discrete filter transfer function in equation (23) are shown in Table 10.

Table 10. Discrete filter transfer function coefficients

| $i$ | $a_i$ | $b_i$ |
|---|---|---|
| 0 | $2.273e - 15$ | 1 |
| 1 | $1.591e - 14$ | $-6.927$ |
| 2 | $4.773e - 14$ | 20.56 |
| 3 | $7.955e - 14$ | $-33.92$ |
| 4 | $7.955e - 14$ | 33.56 |
| 5 | $4.773e - 14$ | $-19.93$ |
| 6 | $1.591e - 14$ | 6.574 |
| 7 | $2.273e - 15$ | $-0.9295$ |

The magnitude response of this 7[th] order low pass Butterworth filter is given as (see Figure 19). Inversing the transfer function, the finite difference equation of the filter is found to be:

$$\begin{aligned} y(n) = {} & 6.927y(n-1) - 20.56y(n-2) + 33.92y(n-3) - 33.56y(n-4) + 19.93y(n-5) \\ & - 6.574y(n-6) + 0.9295y(n-7) + 2.273e - 15x(n) + 1.591e \\ & - 14x(n-1) + 4.773e - 14x(n-2) + 7.955e - 14x(n-3) + 7.955e \\ & - 14x(n-4) + 4.773e - 14x(n-5) + 1.591e - 14x(n-6) + 2.273e \\ & - 15x(n-5) \end{aligned} \quad (24)$$

The transfer function is written to the Simulink model for simulation of real-time filtering of the vehicle dynamics noises. The input parameters of the filter are changed automatically by the MATLAB algorithm as it does for the regression model.

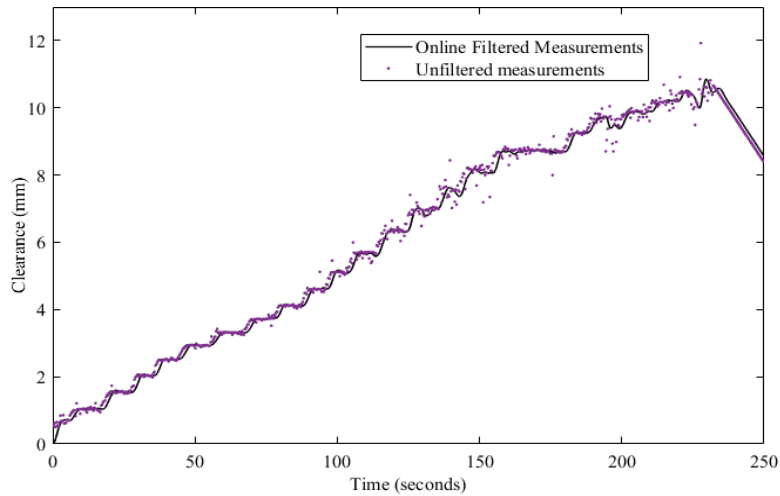

Figure 19. Real-time filtering of wheel flange measurements

Figure 19 shows the results of the simulation of online filtering for emulated wheel flange wear depth measurement. By implementing this filter through simulation, an accuracy of 98.2 % was achieved. However, a slight delay of 10m sec was observed due to computational runtime as the order of the filter increases.

## 6. DISCUSSION OF RESULTS

The developed system integrates both hardware and software components for efficient operation. The hardware includes inductive sensors, temperature sensors, a data acquisition system, and a computer running LabVIEW software. The software implements a machine learning algorithm that processes the acquired data.

The machine learning algorithm is trained using a dataset comprising inductive sensor and temperature readings as input features, with the clearance between the sensor tip and a forward-moving disk serving as the target label. Once trained, the algorithm's predictions are passed through a digital filter to smooth the output and enhance measurement stability.

To ensure long-term accuracy and adaptability, the system periodically collects new sensor data under varying conditions. These updated readings are used for model recalibration, enabling automatic updates to maintain the algorithm's performance over time. This feedback loop ensures the system remains robust and reliable across different operational scenarios.

Equation (3) represents the dynamic machine learning model with a closed form equation to detrmine the learned coefficients and the order of the regression polynomial that satisfies required regression metric in terms of R-square. We set the correlation coefficient of 95%. The model was validated with actuall measurement using standard equipment (comparing the real-time sensor tip- disk surface readings by the system and the micrometer readings (as shown in Figures 8 and 9) and an accuracy of 96.5 % was achieved. This model is built based and sensor output voltage and sensor surrounding temperature and does not includes effects of noise ad rail vehicle dynamics for real-time and Internet of Things measurement. Equation (23) and Equation (24) are the filter algorithms from which their coefficients have been determined. These filter algorithms are destined to be implemented in a real-time device of IoT device to enhance the real-time measurement. In our approach the algorithm in Equation (24) is used to simulate the realtime filtering of the already measured wheel flange wear depth from the machine learning model, without accounting the effects of rail vehicle dynamics. Conversely, equation (24) encompasses these dynamics.

In essence, this indicates that the output of the machine learning algorithm takes into consideration features linked to temperature and sensor output voltage for estimating wheel flange wear depth in real-time. Subsequently, the machine learning output undergoes real-time filtering to mitigate sensor noise and account for vehicle dynamic effects.

The significance of wear depth becomes critical as it approaches failure thresholds. As shown in Figure 19, the MATLAB-based simulated online filtered results reveal a sensor range ($s_r$) between 0 and 10.5 mm. However, this range includes the initial clearance ($s_i$), which must be subtracted to determine the effective wear depth ($s_w$). Thus, the wear depth is calculated as:

$$s_w = s_r - s_i$$

For mainline locomotives, including passenger and freight trains, such as those operating on the Ethio-Djibouti Railway, the typical wheel flange thickness ranges from a maximum of 33 mm to a minimum of 22 mm. This implies an allowable flange thickness reduction due to wear of:

$$s_{W_{max}} = 9\ mm$$

When the flange wear depth ($s_w$) approaches $s_{w_{mx}}$, it is considered critical, as failure is likely to occur. This is due to the insufficient bearing capacity to sustain train lateral forces, which could result in derailment caused by wheel climbing the rail gauge.

To ensure in range readings, the sensor's specification must accommodate a clearance range ($s_r$) greater than $s_{w_{mx}} + s_i$ (the sum of the maximum flange wear depth and the initial clearance between the sensor tip and the wheel flange surface when the wear depth is zero). In the case of the sensor used in this research the minimum $s_i$ would be 1.5 mm.

If the sensor's available range is less than $s_{w_{mx}} + s_i$, critical wear depth will be considered when the sensor reading nears $s_r$. Beyond this point, the readings become unreliable, emphasizing the importance of selecting a sensor range that provides accurate measurements for the entire operational lifecycle.

As shown in Figure 19, the slope of the reading curve represents the rate of wear depth. A steep slope indicates rapid wear of the wheel flange, while a flatter slope suggests minimal or no wear. Accelerated wear may signify that the railcar is operating on irregular track geometry, navigating sharp curves, running on deteriorated track conditions, or traveling on a poorly designed track.

When integrated with internet of things (IoT), this instrument enables real-time monitoring of such conditions, allowing for timely interventions. If excessive wear is detected, appropriate actions—such as stopping the railcar and initiating swift track maintenance—can be undertaken to mitigate potential damage and ensure operational safety.

For this study the digital filter for online filtering was implemented in data acquisition software LABVIEW. The simulations were implemented in the computing software MATLAB to test the online filter performance in terms of runtime and delay. It was observed that real-time filtering in LabVIEW software exhibits higher delay than online simulation in MATLAB/Simulink. It was therefore concluded that hardware implementation using embedded system on chip such as FPGA programmed using hardware descriptive language or any other high level synthesis language such as LabVIEW or MATLAB/SIMULINK, would be a more efficient approach for latency minimization than software abstract implemented real time filter.

# CONCLUSION

The dynamic machine learning algorithm implemented in this study showcases an adaptive regression model that continuously updates as new data is incorporated into the database. The algorithm effectively mitigates measurement drift resulting from temperature effects. To further mitigate effects of sensor time varying behavior, and enhance system response accuracy, frequent addition of more input measurement data from manual tests is essential. This enables the system to self-calibrate and accurately measure wheel flange wear under varying input conditions. As a result, the developed models achieved a correlation coefficient of 95% and an accuracy of 96.5%..

Moreover, the integration of an online digital IIR filter facilitates real-time noise removal from measurements, thereby significantly enhancing system accuracy, as evidenced by real-time filtering simulations. This enhancement is demonstrated by the performance measure of real-time filtering simulations, which achieved 98.2% accuracy. Furthermore, insights obtained from multi-body simulations highlight the influence of higher frequencies induced by train dynamics and track irregularities on measurement accuracy. These insights inform the design parameters of the real-time filter to mitigate such effects. Leveraging this real-time information, seamless integration with railway communication systems is feasible.

In terms of future endeavors, we recommend the hardware implementation of the real-time IIR filter to enhance system performance, particularly in response time. Additionally, considering the behavior of wheelsets and locomotive bogie displacements on curved tracks is vital for specifying an optimal design for the sensor support structure mechanism. This holistic approach promises to refine the accuracy and reliability of the system for enhanced railway safety and maintenance operations.

# ACKNOWLEDGMENTS

The authors acknowledge the World Bank for the support provided in this research through grant to the African Railway Center of Excellence. The authors acknowledge the National Instruments Company for providing the LABVIEW and DAQ software trial licenses. The authors acknowledge the support provided by Computational Mechanics ltd for providing us the free six months license of Universal Mechanism software.

# FUNDING INFORMATION
This research was partially funded by Addis Ababa University, Ethiopia.

# AUTHOR CONTRIBUTIONS
The authors confirm contribution to the paper as follows: study conception and design: C. Nkundineza, J. Ndodana; data collection: J.N. Njaji, Samrawit Abubeker; analysis and interpretation of results: J. N. Njaji, C. Nkundineza. Samrawit Abubeker, Damien Hanyurwimfura, Omar Gatera; Draft manuscript preparation: J.N. Njaji, C. Nkundineza. Omar Gatera. All authors reviewed the results and approved the final version of this manuscript.

# CONFLICT OF INTEREST STATEMENT

Authors state there is no conflict of interest

# DATA AVAILABILITY STATEMENT

The data generated and analyzed during this study are available from the corresponding author upon reasonable request.

**Appendices**

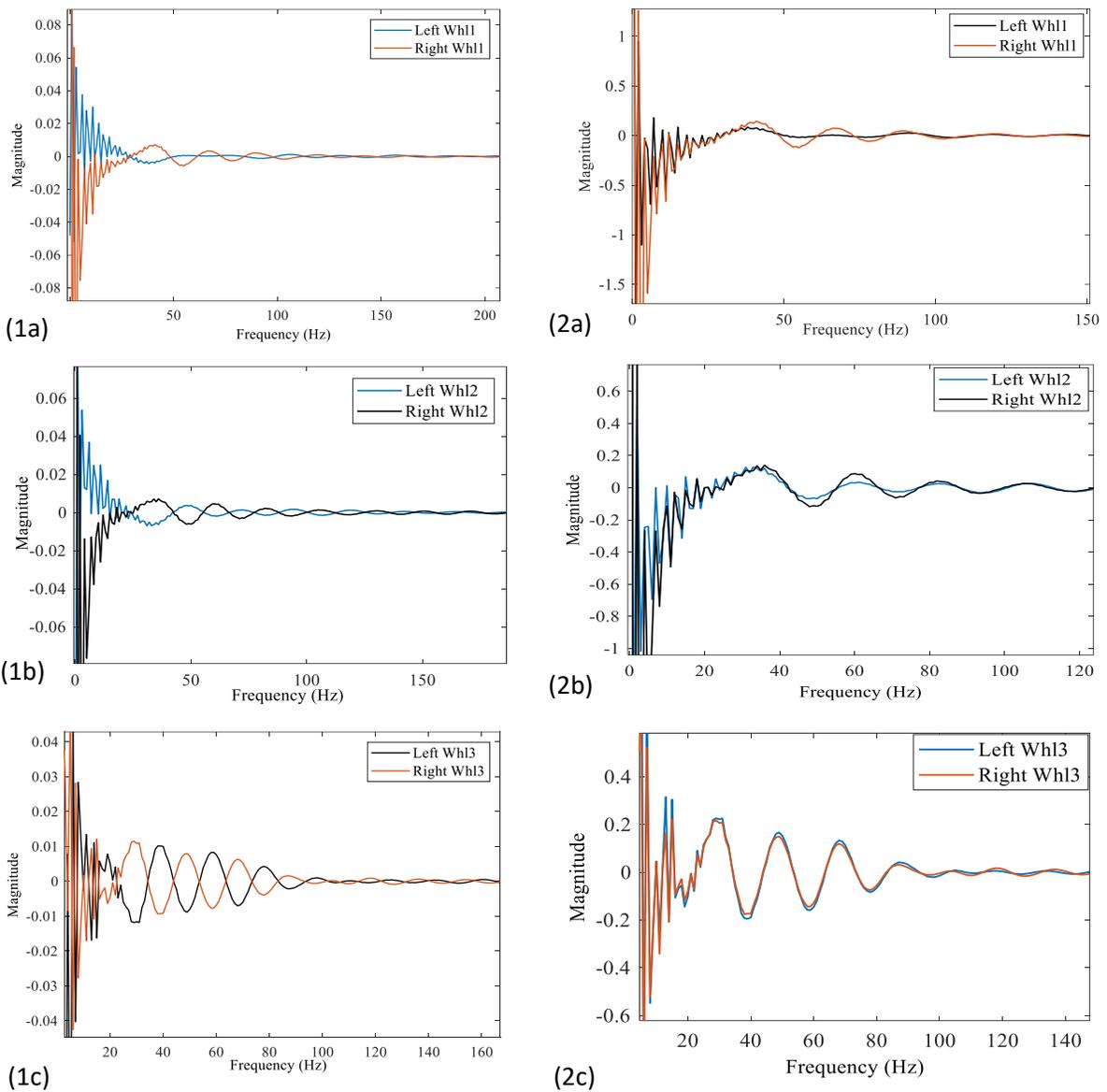

Figure A1. FFT analysis of vertical, (1a), (1b), (1c); and fft analysis of lateral displacement, (2a), (2b), (2c)

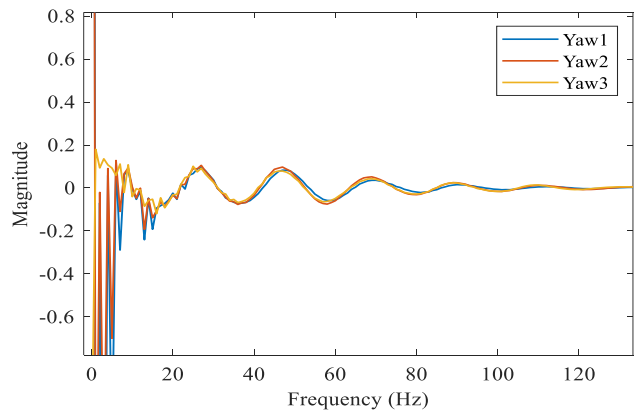

Figure A2. FFT analysis of the yaw motions of wheelsets in bogie 1

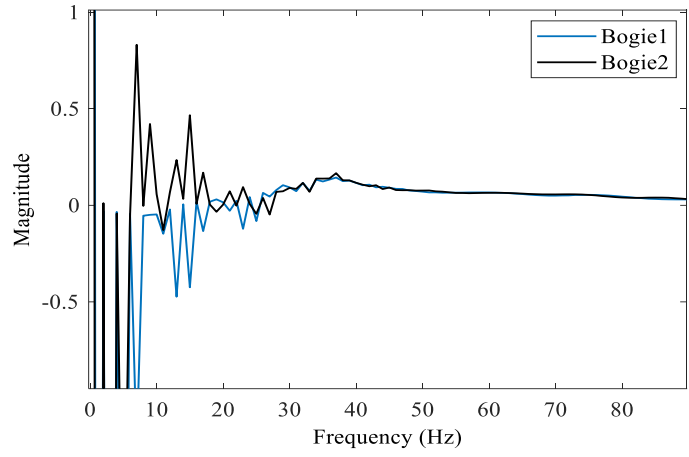

Figure A3. FFT analysis of the vertical motions of the bogies